\documentclass[conference]{IEEEtran}
\IEEEoverridecommandlockouts
\usepackage{cite}
\usepackage{amsmath,amssymb,amsfonts}
\usepackage{algorithmic}
\usepackage{graphicx}
\usepackage{textcomp}
\usepackage{xcolor}
\usepackage{array}
\usepackage{soul}
\usepackage[ruled,vlined]{algorithm2e}
\usepackage{booktabs,ragged2e}
\usepackage[flushleft]{threeparttable}
\usepackage{bm}
\usepackage[letterpaper, left=0.68in, right=0.68in, bottom=1in, top=0.72in]{geometry}
\def\BibTeX{{\rm B\kern-.05em{\sc i\kern-.025em b}\kern-.08em
    T\kern-.1667em\lower.7ex\hbox{E}\kern-.125emX}}
\begin{document}

\title{FedGreen: Federated Learning with Fine-Grained Gradient Compression for Green \\Mobile Edge Computing }

\author{\IEEEauthorblockN{Peichun Li\IEEEauthorrefmark{1},
Xumin Huang\IEEEauthorrefmark{1}, Miao Pan\IEEEauthorrefmark{2} and
Rong Yu\IEEEauthorrefmark{1}}
\IEEEauthorblockA{\IEEEauthorrefmark{1}School of Automation, Guangdong University of Technology, Guangzhou, China\\
\IEEEauthorrefmark{2}Department of Electrical and Computer Engineering, University of Houston, Houston, USA\\
Email: peichun@mail2.gdut.edu.cn,
huangxu\_min@163.com,
mpan2@uh.edu,
yurong@ieee.org}}

\maketitle

\begin{abstract}
Federated learning (FL) enables devices in mobile edge computing (MEC) to collaboratively train a shared model without revealing the local data. Gradient compression could be applied to FL to alleviate the communication overheads but the existing schemes still face challenges. To deploy green MEC, we propose FedGreen, which enhances the original FL with fine-grained gradient compression to control the total energy consumption of the devices. Specifically, we introduce the relevant operations including device-side gradient reduction and server-side element-wise aggregation to facilitate the gradient compression in FL. According to a public dataset, we evaluate the contributions of the compressed local gradients with respect to different compression ratios. Furthermore, we investigate a learning accuracy-energy efficiency tradeoff problem and the optimal compression ratio and computing frequency are derived for each device. Experimental results show that given the 80\% test accuracy requirement, compared with the baseline schemes, FedGreen reduces at least 32\% of the total energy consumption of the devices.

\end{abstract}


\section{Introduction}
Federated learning (FL) is a promising distributed machine learning framework that enables multiple devices to jointly train a shared model by their private datasets while preserving the training data privacy \cite{mcmahan2017communication}. In FL, a parameter server with the central position distributes an initialized learning model to the devices. Each device trains the model by the local dataset and submits the local gradients to the parameter server. All local gradients are aggregated to update the global model. Then the updated global model is sent to each device to perform a new local model training task. The iterative training procedure is repeated until convergence. Recently, the emerged computing paradigm named mobile edge computing (MEC) is applied to facilitate the execution of FL \cite{mao2017survey}. Massive edge devices in MEC posse versatile sensors to collect raw data and have under-utilized resources to execute the FL algorithm. Many research efforts have been devoted to optimizing the performance of FL in MEC.

Researchers have integrated gradient compression into FL to compress the local gradients of the devices and decrease the number of bits transmitted to the parameter server. For example, some less important local gradients were clipped based on the magnitude \cite{shi2019convergence} and let a small number of bits represent the gradient values \cite{8889996}. Similarly, a universal vector quantization scheme was studied in \cite{shlezinger2020uveqfed}.  But these methods neglected that different devices have different channel states, computing capabilities and energy consumption rates such that they could require different compression ratios to match with their energy states.  In addition to the uniform gradient compression, device scheduling is introduced to provide unified management for all devices according to diverse optimization goals \cite{nishio2019client, 9142401, 8964354}. The methods select specific devices to perform local training tasks and this could accelerate the training procedure of FL to a degree. But the methods directly limit the amount of training data and cause the unbalanced usage of all devices' data.

Toward green deployment of MEC, FL with gradient compression still faces great challenges. To execute the FL algorithm, local model training requires each device to consume a certain number of computation resources \cite{li2021talk, shi2021towards}. At the same time, wireless bandwidth is necessitated since learning in a decentralized manner takes hundreds of communication rounds until convergence. But devices in MEC are generally battery-limited. For a green MEC system, the total energy consumption of the devices should be controlled to create energy savings and avoid battery degradation. In turn, a variety of devices in MEC may have heterogeneous resources in terms of computation, communication, and power \cite{yu2021}. In FL with gradient compression, the computing frequency and compression ratio of each device should be optimized to match the hardware configuration and channel status. 

To promote the FL with gradient compression, we adopt different compression ratios for different devices in MEC and study a learning accuracy-energy efficiency tradeoff problem. We present a comprehensive scheme called by FedGreen, which enhances the original FL with fine-grained gradient compression to achieve green MEC. Specifically, we first present a basic method that enables each device to choose a specific ratio to compress the local gradient after local model training.  As a consequence, a device can switch to a small compression ratio to report more accurate gradient information in the resource-sufficient state, and a large one to decrease the communication overheads and save energy in the resource-deficient state.  We further study how to derive an acceptable compression ratio for each device.  According to a public dataset,  the quantitative relationship between gradient compression ratio and global model accuracy is formulated. To balance the accuracy performance and total energy consumption of the devices, we investigate the learning accuracy-energy efficiency tradeoff problem for FL with gradient compression, and jointly optimize the compression ratios and computing frequency of the devices. 
Extensive experimental results are provided to validate the efficiency and effectiveness of the proposed scheme.

The main contributions of the paper are summarized as follows.
\begin{itemize}
	\item We design a fine-grained gradient compression method by combining device-side gradient reduction and server-side element-wise aggregation. Based on current techniques, FedGreen enables different devices to compress the local gradients on demand, according to the energy states.
	
	\item We present a learning accuracy-energy efficiency tradeoff problem for FL with gradient compression.  The compression ratio and computing frequency of each device are jointly optimized to ensure the algorithm performance of FL while reducing the total energy consumption.
	
	\item We conduct experiments to validate the overall performance of FedGreen. Compared with the baseline schemes of FL, FedGreen saves energy on the devices and achieves fine-grained gradient compression for green MEC.
\end{itemize}

The rest of this paper is organized as follows. 
We describe the fine-grained gradient compression method in Section II. Section III discusses the learning accuracy-energy efficiency tradeoff problem and its theoretical analysis. Experiment evaluation of our framework is shown in Section IV. Finally, Section V concludes this paper.

\section{Fine-Grained Gradient Compression}

\subsection{Device-side Gradient Compression}
Without loss of generality, we take the two-dimension convolution layer as an example, and consider the layer-wise gradient compression. The three-stage gradient compression consists of sparsification, quantization and encoding. Let $\bm{v}\in \mathbb{R}^N$ denote the original gradients before compression and $N=C_{out}\times C_{in} \times K \times K$, where $C_{out}$, $C_{in}$ and $K$ are the \#output channels, \#input channels and kernel size, respectively. Here, we use 32 bits to represent a float number.

\begin{figure}
\centering
 \includegraphics[width=0.48\textwidth]{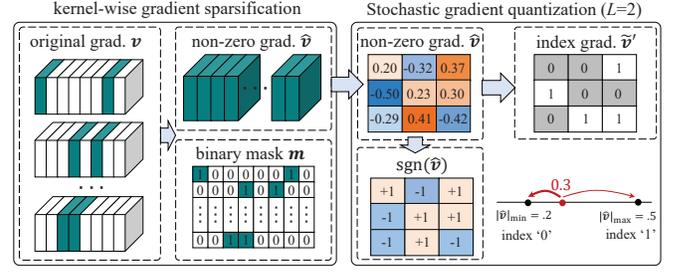}
 \caption{Gradient sparsification and quantization.}\label{spar-quant}
 \vspace{-0.45cm}
\end{figure}

{\bf Kernel-wise gradient sparsification}.
We define the \emph{kernel} with shape of $K \times K$ as the basic unit of the gradient sparsification.  As shown in the left of Fig.~\ref{spar-quant}, we calculate the L2 norm of each kernel in one layer. Given a pruning rate of $\rho \in [0,1)$, we zero-out the first $\lfloor \rho C_{out}C_{in} \rfloor$ kernels with smallest norm. Let $\bm{v}_s = \bm{v} \odot \bm{m}, \bm{v}_s \in \mathbb{R}^N $ represent the sparse gradient after pruning, where $\bm{m}$ is the binary mask with shape of $C_{out}\times C_{in}$ and $\odot$ is the Hadamard product with broadcasting. Furthermore, let $\hat{\bm{v}}$ represent the non-zero entries in $\bm{v}_s$, and $\hat{\bm{v}} \in \mathbb{R}^M$, where $M=\lceil {(1-\rho)C_{out}C_{in}} \rceil K^2 $. Given a mask $\bm{m}$ and the non-zero gradient $\hat{\bm{v}}$, we can obtain $\bm{v}_s$ by $\bm{v}_s = {\cal R}(\hat{\bm{v}}, \bm{m})$, where ${\cal R}$ is the function that reconstructs the sparse gradient from the dense one with respect to $\bm{m}$.  Compared with the traditional filter-wise method that zeroes out a whole filter to compress the gradient, kernel-wise sparsification achieves fine-grained pruning while maintaining a small mask size.

\textbf{Lemma 1.} For any sparse gradient ${\bm{v}}_s$ obtained from $\bm{v} \in \mathbb{R}^N$ and $\rho \in [0,1)$ by kernel-wise sparsification, we have ${\left\| {\bm{v} - {\bm{v}}_s} \right\|_2^2} \leq {\rho\left\| \bm{v} \right\|_2^2}$.

{\bf Stochastic gradient quantization}. Motivated by QSGD in \cite{alistarh2017qsgd}, we propose a reinforced stochastic quantization scheme for the pruned gradient $\hat{\bm{v}}$. Let $\left| {\hat{\bm{v}}} \right|_{\min}$ and $\left| {\hat{\bm{v}}} \right|_{\max}$ be the minimum and maximum value of $\left| {\hat{\bm{v}}} \right|$, and $\Delta = {\left| {\hat{\bm{v}}} \right|_{\max}} - \left| {\hat{\bm{v}}} \right|_{\min} $. Let $j$ index the entries in $\hat{\bm{v}}$, and $L$ be the number of quantization levels. We can quantize any non-zero scalar $\hat{v}^j \in \hat{\bm{v}}$ by
\begin{equation}
{\cal Q}(\hat{v}_j, L) = \textrm{sgn} (\hat{v}^j) \cdot\left[ \delta(\left| {\hat{v}}^j \right|-\left| {\hat{\bm{v}}} \right|_{\min};L)+\left| {\hat{\bm{v}}} \right|_{\min}\right],
\end{equation}
where $\textrm{sgn}(\hat{v}^j) \in\{-1, +1\}$ denote the sign of $\hat{v}^j$, and ${\tilde x} = \delta(x;L)$ is the stochastic quantization function that maps $x\in [0, \Delta]$ to ${\tilde x} \in \{0, \frac{\Delta}{L-1}, \frac{2\Delta}{L-1}, \cdots, \Delta\}$. Let $l\in\{0, 1, \cdots, L-1\}$ be an integer such that $x\in[\frac{\Delta l}{L},\frac{\Delta (l+1)}{L}]$. Hence, $[\frac{\Delta l}{L},\frac{\Delta (l+1)}{L}]$ is the quantization interval of $x$. Then, we have
\begin{equation}
\delta(x;L)=
\left\{ {\begin{array}{*{20}{c}}
\Delta (l+1)/L&\textrm{with probability } {xL/\Delta-l}\\
\Delta l/L&\textrm{otherwise}.
\end{array}} \right.
\end{equation}
After applying ${\cal Q}(\hat{v}^j, L)$ for all $\hat{v}^j \in \hat{\bm{v}}$, we obtain the quantizated gradient $\tilde{\bm{v}}$. Naturally, $\tilde{\bm{v}}$ can be represented by a tuple $(\tilde{\bm{v}}^\prime, \textrm{sgn}(\hat{\bm{v}}), \left| {\hat{\bm{v}}} \right|_{\min}, \left| {\hat{\bm{v}}} \right|_{\max})$, where $\tilde{\bm{v}}^\prime$ is the index gradient with each entry of $\log _2 L$ bits. Given a mask $\bm{m}$ and its quantized gradient $\tilde{\bm{v}}$, we can obtain the sparse form of quantized gradient $\bm{v}_q = {\cal R}(\tilde{\bm{v}}, \bm{m})$. An example of quantization process is provided in right of Fig.~\ref{spar-quant}.

\textbf{Lemma 2.} For any quantizated gradient $\bm{v}_q \in \mathbb{R}^N$ computed from pruned gradient $\bm{v}_s = {\cal R}(\hat{\bm{v}}, \bm{m}), \hat{\bm{v}} \in \mathbb{R}^M$ by the above scheme with $L$ levels, we have ${\left\| {\bm{v}_q - \bm{v}_s} \right\|_2} \leq M\Delta /(L-1)$.

{\bf Lossless encoding}. Now, we obtain a binary mask $\bm{m}$ and a tuple $(\tilde{\bm{v}}^\prime, \textrm{sgn}(\hat{\bm{v}}), \left| {\hat{\bm{v}}} \right|_{\min}, \left| {\hat{\bm{v}}} \right|_{\max})$. Since $\bm{m}$ is sparse, we utilize compressed sparse row (CSR) format to represent $\bm{m}$ and obtain ${\cal E}_{\textrm{csr}}(\bm{m})$. Furthermore, due to the statistical characteristics of $\tilde{\bm{v}}^\prime$ that smaller indices are more frequent, we apply Huffman coding to reduce the data size and get ${\cal E}_{\textrm{H}}(\tilde{\bm{v}}^\prime)$. Finally, we get an encoded tuple with five 
parts $({\cal E}_{\textrm{csr}}(\bm{m}), {\cal E}_{\textrm{H}}(\tilde{\bm{v}}^\prime), \textrm{sgn}(\hat{\bm{v}}), \left| {\hat{\bm{v}}} \right|_{\min}, \left| {\hat{\bm{v}}} \right|_{\max})$.

\textbf{Lemma 3.} Given any convolution gradient $\bm{v}\in \mathbb{R}^N$, by combining the above compression schemes with pruning rate $\rho$ and quantization levels $L$, the number of bits to communicate $({\cal E}_{\textrm{csr}}(\bm{m}), {\cal E}_{\textrm{H}}(\tilde{\bm{v}}^\prime), \textrm{sgn}(\hat{\bm{v}}), \left| {\hat{\bm{v}}} \right|_{\min}, \left| {\hat{\bm{v}}} \right|_{\max})$ is upper bounded by
\begin{equation} \label{upperbound}
C_{out}C_{in} + 
\lceil {(1-\rho)C_{out}C_{in}} \rceil K^2(1+\log _2 L) + 64. 
\end{equation}

Specifically, we use a fixed $L=8$ for convolution layer and $L=4$ for fully connected layer during the implementation and the compression ratio is only determined by $\rho$. Naturally, according to Lemma 3, there is a near-linear relationship between $\rho$ and the size of compressed gradient. We can directly acquire the gradient pruning rate $\rho$ for a given compression ratio. 
Unlike previous gradient compression methods\cite{alistarh2017qsgd, shlezinger2020uveqfed} that only reduce the local gradient size with a predefined set of compression ratios, the proposed method can perform fined-grained gradient compression in large range sizes.
Note that the computation cost of local gradient compression is negligible compared to that of local model training.

\subsection{Server-side Element-wise aggregation}
There exist $I$ devices that collaboratively a shared model, and we utilize ${\cal I} = \{1, 2, \cdots, I\}$ to denote the device set. After collecting the compressed gradients uploaded from different devices, the parameter server first decodes the compressed gradients and obtains $\{\bm{v}_{q,i}, \forall i \in {\cal I}\}$. Then the parameter server is responsible to compute the global gradient $\overline{\bm{v}}$.

\begin{figure}
\centering
 \includegraphics[width=0.43\textwidth]{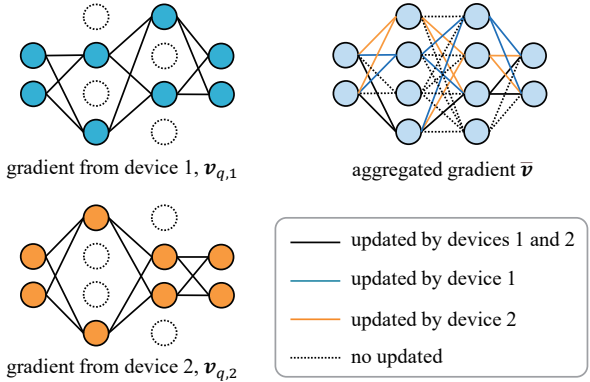}
 \caption{Gradient aggregation.}\label{agg-fig}
 \vspace{-0.5cm}
\end{figure}

Let $\bm{v}_{q,i}=\{{v}_{q,i}^k\}~(k=1,2,\cdots, N)$ represent the compressed gradient uploaded from device $i$, and $\bm{m}_i=\{m_i^k\}$ be its corresponding mask. The aggregated gradient can be expressed in an element-wise manner by $\overline{\bm{v}}=\{\overline{v}^k\}~(k=1,2,\cdots, N)$. The $k$-th entry $\overline{v}^k$ is calculated by
\begin{equation}\label{agg-eqn}
\overline{v}^k = \left\{ {\begin{array}{*{20}{c}}
{\frac{1}{{\sum\nolimits_i {m_i^k{D_i}} }}\sum\nolimits_i {v_{q,i}^km_i^k{D_i}} ,}&{\textrm{if}~\sum\nolimits_i {m_i^k{D_i}}  > 0}\\
{0,}&{\textrm{otherwise}}
\end{array}} \right.,
\end{equation}
where $D_i$ is the number of training data of device $i$.

We take an example to compute $\overline{\bm{v}}$ in the multilayer perceptron case, as illustrated by Fig.~\ref{agg-fig}. We consider a simple application scenario of FL consisting of two devices. Here, we paint all gradient values of $\overline{\bm{v}}$ with 2 colors. Based on Eqn.~(\ref{agg-eqn}), the connections in blue are only updated by device 1, the connections in orange are only updated by device 2, and the connections in black are updated by both devices 1 and 2, etc. Note that the method is straightforward to be extended to the case of convolution layer.

\subsection{Compressed Gradient Information}
Referring to the tradeoff between model accuracy and communication overhead in \cite{konevcny2016federated}, we infer that a high compression ratio in gradient compression leads to the deterioration of global model accuracy. In the following, we study the quantitative relationship between the compression ratio and global model accuracy.

\begin{figure}
\centering
 \includegraphics[width=0.42\textwidth]{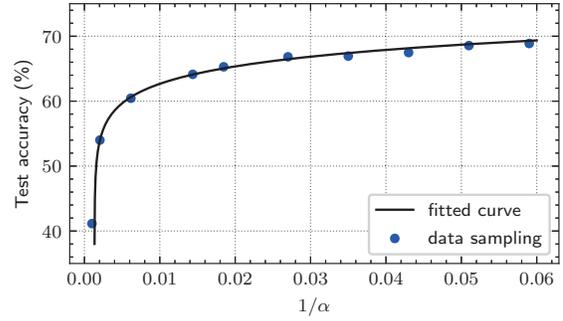}
 \caption{Test accuracy on proxy dataset with respect to $1/\alpha$}\label{acc-ratio}
 \vspace{-0.5cm}
\end{figure}

We refer to the parameter fitting method in the previous work \cite{8963610} and conduct experiments of gradient compression to measure different global model accuracy with respect to different compression ratios $\alpha$ of the devices. To this end, a naive fitting approach is to directly acquire the corresponding global model accuracy by enumerating a set of compression ratios in FL. However, the cost incurred by multiple times of decentralized training may overtake the gain of the parameter fitting itself. Alternatively, we propose to explore the prior knowledge of parameter fitting on a \emph{proxy} task with \emph{public} dataset, and then transfer it to the target task with decentralized dataset. Note that the idea of \emph{proxy} dataset is widely used in the study of neural architecture search \cite{zoph2018learning}. The overall parameter fitting experiments are performed in an offline manner, and the prior knowledge of this one-time fitting can be transferred into \emph{many} FL tasks.

We adopt the CINIC \cite{darlow2018cinic} as the proxy dataset for the parameter fitting experiment. The details of the hyperparameter settings are shown in Section IV.
In the experiments, we apply the control variate method. Given a unified compression ratio of the device, we evaluate the global model accuracy as $F(\alpha)$ after gradient compression. According to the previous work of model compression \cite{han2015deep}, there is a logarithmic relationship between the inversion of compression ratio and the compressed model accuracy. We observe that this relationship is also achieved in FL with gradient compression. Hence, we are motivated to formulate the global model accuracy $F(\alpha)$ by
\begin{equation}\label{fit-eqn}
F(\alpha) = {\kappa _1}\log_2 ({\kappa _2}/{\alpha} -{\kappa _3}) + {\kappa _4},
\end{equation}
where parameters $\kappa_1$, $\kappa_2$, $\kappa_3$ and $\kappa_4$ are experimentally fitted to measure the training performance with the given unified compression ratio $\alpha$. The experimental results are presented in Fig.~\ref{acc-ratio}. We obtain the constant parameters ${\kappa _1}=0.024, {\kappa _2}=19.221, {\kappa _3}=2.561, {\kappa _4}=0.609$. With the decrease of $\alpha$, more accurate gradient information is collected from the compressed local gradients, which is helpful to improve the global model accuracy. For example, when $\alpha$ is small enough (e.g., $\alpha\leq 25$),  the compressed local gradient is able to reveal sufficient and accurate gradient information and at this time, the global model accuracy is almost identical to that of the conventional FL algorithm. Therefore, we consider that for a single device, a lower compression ratio $\alpha$ could also give rise to more accurate gradient information and vice versa.

\section{Problem Formulation and Solution}
\subsection{Learning Accuracy-Energy Efficiency Tradeoff Problem}
To study FedGreen, we consider an application scenario of FL including a parameter server co-located with a base station and $I$ devices. Let $f_{i}$ and $r_{i}$ denote the computing frequency (CPU cycles/second) and uplink data rate (bps) of device $i$, respectively. Given the model structure, the original weight and gradient size $S$ and computing workloads per training sample $W$ are easy to calculate. The number of local epochs $n$ for model training could be set empirically. Similar to \cite{9264742}, we pay attention to the total latency of local model training and gradient uploading in each communication round of the global model training. Specifically, with the instruction of the parameter server, each device independently trains the published global model with the local dataset. The local training time is ${nWD_i}/{f_{i}}$. During the gradient uploading process, due to a compression ratio $\alpha_{i}$, gradient uploading time of device $i$ is consumed by $S/{(\alpha_{i}r_{i})}$. For multiple access of the devices in the uplink data transmission, we consider device $i$ communicates with the base station via the frequency domain multiple access technology. For device $i$,  the uplink data rate $r_i$ is calculated by
\begin{equation}
	r_i={{b_i}\log_2 (1 + \frac{{p_i{{\left| {{h_i}} \right|}^2}}}{{{N_0}{b_i}}})},
\end{equation}
where $b_i$ and $p_i$ indicate the available bandwidth and transmitter power of device $i$ respectively, $h_i$ represents channel gain between the device and base station, and $N_0$ indicates the noise power-spectral-density. In addition to the time consumption, the amount of energy consumed for local model training is $\varepsilon_i f_i^2 nD_iW_i $, where $\varepsilon_i$ is an energy coefficient of the device. As introduced by \cite{huang2021fedparking}, $\varepsilon_i$ indicates the effective switched capacitance relying on the chip architecture.

In FL with gradient compression, a low compression ratio $\alpha_{i}$ reduces the data size of the local gradient and causes less communication overheads to device $i$ in the uplink data transmission. But this leads to less accurate gradient information, the global model accuracy will be degraded to a degree.  Besides, similar to the conventional FL algorithm, we consider the influence of  the amount of local training data $D_i$ when evaluating the contribution of the gradient information submitted by device $i$.  We utilize $D_i/{\cal D}$ as a weighting factor of device $i$, where ${\cal D} = \sum\nolimits_i {{D_i}}$. Ultimately, we measure the overall contribution of all the compressed local gradients from the devices by
\begin{equation}
	{{\mathcal F}({{\{ {\alpha _i}\} }_{1 \le i \le I}})} = \frac{1}{{\mathcal D}}\sum\limits_{1\le i\le I} {{D_i}F({\alpha _i})},
\end{equation}
where $F(\alpha_i)$ is computed by Eqn.~(\ref{fit-eqn}) to roughly measure the training performance of device $i$ after the device compresses its local gradient by ratio $\alpha _i$. Until now, we introduce $\mathcal{F}(\{\alpha_{i},\forall i\})$ as a new performance metric to evaluate the learning performance of FL with gradient compression.

Considering the learning performance of FL and total energy consumption of all the devices, there exists a tradeoff problem in FL with gradient compression. The goal function can be expressed by
\begin{equation}\label{objective}
\mathcal G={{\cal F}({{\{ {\alpha _i}\} }_{1 \le i \le I}})}- \varpi J\sum\nolimits_{i = 1}^I {(\frac{{{p_i}S}}{{{\alpha _i}{r_i}}} + {\varepsilon _i}f_i^2{n}{D_i}{W})},
\end{equation}
where $J$ is the predefined number of global iterations and $\varpi$ is a presetting weighting factor. To achieve the goal, we jointly optimize the compression ratio $\alpha_i$ and computing frequency $f_i$ of each device $i, 1\le i \le I$. At the same time,  there are essential constraints for the tradeoff problem. The compression ratio $\alpha_{i}$ is equal to or larger that 1 and $f_i^{\max}$ is an upper limit of  $f_i$.  Moreover, a parameter server-defined latency constraint should be satisfied for each device. Here, $T^{\max}$ is the training delay requirement of a single global iteration for each device. Finally,  we summarize the whole problem with necessary constraints as follows.
\begin{equation}\label{problem}
	\begin{array}{*{20}{l}}
		{({\textrm{P1}}):}&\max~{{\mathcal G}} \\
		{{\textrm{subject to: }}}&{\alpha_i \ge 1, \forall i,}\\
		{}&{0 < f_i \le f_i^{\max }, \forall i,}\\
		{}&{{\frac{S}{{{\alpha _i}{r_i}}} + \frac{{{n}{D_i}W}}{f_i} \le {T^{\max }},\forall i}},\\
		{{\textrm{variables}}:}&{\alpha_i, f_i, \forall i}
	\end{array}
\end{equation}

For efficiency guarantee, the above optimization problem is solved by each device in a decentralized manner. Specifically, finding the optimal solution $\{f_i^*, \alpha_i^*\}$ for device $i$ only requires its own hardware states and channel state information. Hence, each device can dynamically update its training strategy to cope with the time-varying environment during the training period.
After solving the problem, each device utilizes a suitable computation frequency to perform the local training task and afterward compress the local gradient with a specific ratio. 
In this paper, we design FedGreen to reduce the total energy consumption of all the devices in the goal function and also consider a latency constraint for each device. Our scheme is beneficial to achieve FL with fine-grained gradient compression for green MEC.


\subsection{Solution}
To tackle the above optimization problem, we first pay attention to the bottom computing resource allocation problem. With the decisions of gradient compression $\left\{\alpha_i, \forall i\right\}$,   the subproblem that only involves the decision variables $\left\{f_i, \forall i\right\}$ is formulated to minimize the total energy consumption cost of the devices, which is expressed as follows
\begin{equation}
	\begin{array}{*{20}{l}}
		{({\textrm{P1-Bottom}}):}&{\min~{\sum\limits_{1 \le i \le I} {(\frac{{{p_i}S}}{{{\alpha _i}{r_i}}} + {\varepsilon _i}f_i^2{n}{D_i}{W_i})} } } \\
		{{\textrm{subject to: }}}&{0 < f_i \le f_i^{\max }, \forall i,}\\
		{}&{{\frac{S}{{{\alpha _i}{r_i}}} + \frac{{{n}{D_i}W}}{f} \le {T^{\max }},\forall i}}\\
		{{\textrm{variable: }}}&{f_i, \forall i}
	\end{array}
\end{equation}
Since the total energy consumption increases with the increase of $f_i$, we realize that if $\alpha_i$ is confirmed for device $i$, $f_i$ is solved according to the time delay constraint and the upper limit. Hence, the solution of $f_i$ is
\begin{equation}
	f_i^* = \min (\frac{{{n}{D_i}W}}{{{T^{\max }} - \frac{S}{{{\alpha _i}{r_i}}}}},f_i^{\max }). \label{optf}
\end{equation}

Note that when $f_i^{\max}$ is large enough, $f_i$ is straightforwardly solved according to the equalized the time delay constraint. We utilize an intermediate variable $\beta_i \in [0,1]$ to suppose that for device $i$,
\begin{equation}
	\left\{ \begin{array}{l}
		\frac{S}{{{\alpha _i}{r_i}}} = \beta_i {T^{\max }},\\
		\frac{{{n}{D_i}W}}{{{f_i}}} = (1 - \beta_i ){T^{\max }}.
	\end{array} \right. \label{afb}
\end{equation}
Considering that $f_i \le f_i^{\max}$, a lower limit of $\beta_i$ is required by 
	\begin{equation}
	{\beta_i^{\min }} = 1 - \frac{{{n}{D_i}W}}{{f_i^{\max }{T^{\max }}}}.
	\end{equation}
We derive the partial derivatives of $\mathcal {G}$ with respect to $\beta_i$,
\begin{equation}
	\left\{ {\begin{array}{*{20}{l}}
			{\frac{{\partial {\cal G}}}{{\partial {\beta _i}}} = \frac{{{r_i}{T^{\max }}}}{S}\frac{{{D_i}{\kappa _1}{\kappa _2}}}{{{\cal D}({\kappa _2}{\lambda _i} - {\kappa _3})\ln 2}}}\\
			{ - \varpi J({p_i}{T^{\max }} + 2{\varepsilon _i}\frac{{{n^3}{D_i}^3{W^3}}}{{{{({T^{\max }})}^2}{{(1 - \beta )}^3}}})},\\
			\begin{array}{l}
				\frac{{{\partial ^2}{\cal G}}}{{\partial \beta _i^2}} =  - {(\frac{{{r_i}{T^{\max }}}}{S})^2}\frac{{{D_i}{\kappa _1}\kappa _2^2}}{{{\cal D}{{({\kappa _2}{\lambda _i} - {\kappa _3})}^2}\ln 2}}\\
				- 6\varpi J\frac{{{\varepsilon _i}{n^3}{D_i}^3{W^3}}}{{{{({T^{\max }})}^2}{{(1 - \beta )}^4}}} < 0,
			\end{array}\\
	\end{array}} \right.
\end{equation}
where ${\lambda_i} = 1/{\alpha_i}$. Clearly, $\mathcal{G}$ is concave on $\beta_i$ and we solve the optimal solution of $\beta_i$ based on the first-order optimality condition $\partial {\mathcal{G}}/\partial {\beta_i}=0$. But it is difficult to directly solve $\beta_i$. Alternatively, we apply the binary search method to seek an approximate solution of $\beta_i$ within the range $[\beta _i^{\min },1]$. Finally, $\alpha_i$ and $f_i$ are solved by substituting the approximate solution of $\beta_i$ into Eqn.~(\ref{afb}). 


\section{Performance Evaluation}
\subsection{Experiment Setting}
We consider the application of FL for image classification on the CIFAR-10 dataset, with 16 mobile devices. The hyperparameters for the FL algorithm are shown as follows: local epoch 1, batch size 64, learning rate 0.05, the number of global iterations 300, and decay rate per round 0.996 by default. For the IID data setting, we shuffle the training samples and uniformly dividing them to all the devices. For the non-IID setting, we consider heterogeneous partition with distribution of $\textbf{\textrm{p}}_c\sim\textrm{Dir}_{c,i}(0.5)$ and allocate a proportion $\textbf{\textrm{p}}_{c,i}$ of the training samples of class $c$ to device $i$. We conduct the experiments on VGG-9 model \cite{simonyan2014very}. The original 
gradient size and computing workloads per training sample are empirically measured as $S=111.7$ Mb and $W=0.98$ megacycles, respectively. The parameter settings for the hardware configuration and channel status of the devices are shown as follows: energy coefficient $\varepsilon _i \sim U[5\times 10^{-27}, 1\times 10^{-26}]$, computing frequency $f_{i}^{\max} \sim U[1.5, 4]$ GHz, power-spectral-density $N_0=-114$ dBm, available bandwidth $b_i \sim U[0.8, 5]$ MHz. We set the weighting factor $\varpi = 1\times 10^{-4}$ and $T_{\max}=100$ seconds by default.

\subsection{Performance Evaluation}

\begin{figure*}[t]\centering
  \includegraphics[width=0.98\textwidth]{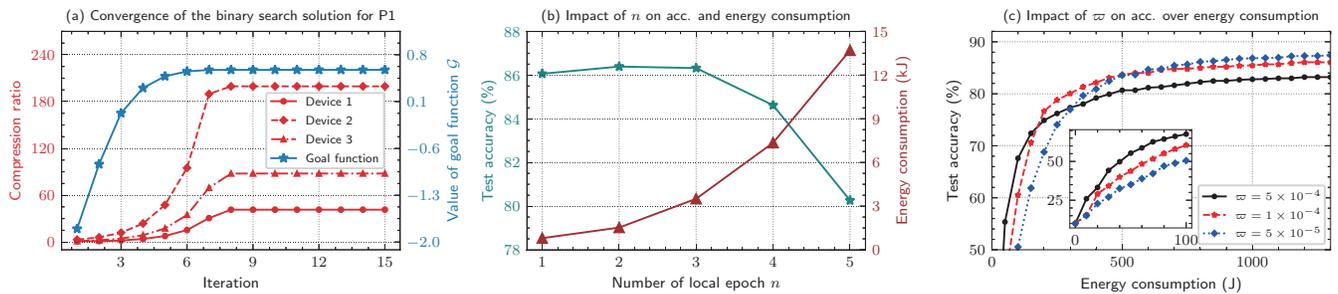}
  \caption{The convergence of binary search solution and the impact of different parameters on the overall performance of FedGreen.}\label{hyper}
  \vspace{-0.2cm}
\end{figure*}

{\bf Convergence of the binary search solution}. We first show that the binary search method can converge finally and achieve the approximately optimal solution for problem P1. Fig.~\ref{hyper}(a) shows the evolution of the compression strategies of three randomly selected devices and the convergence of goal function. In our scheme, a device with lower computation or communication capacity is suggested to adopt a large compression ratio in gradient compression, and vice versa.

{\bf Impact of local epoch $n$}. A larger $n$ encourages each device to perform more iterations of the local model training. But it may lead to the divergence of the local gradients. As shown in Fig.~\ref{hyper}(b), we observe a degradation of the final test accuracy when $n>4$, which matches with the existing study in \cite{Wang2020Federated}. Besides, with the increase of $n$, the energy consumption of local model training increases drastically. Hence, it is recommended to use a moderate local epoch (e.g., $n<5$) to save energy and avoid the gradient divergence.

{\bf Impact of weighting factor $\varpi$}. We discuss the impact of the weighting factor on the performance of our scheme. Fig.~\ref{hyper}(c) shows that we conduct experiments under IID data setting with respect to $\varpi$. A large value of $\varpi$ means that the parameter server would like to reduce the total energy consumption of the devices. The convergence accuracy of the FL algorithm could be a sacrifice at this time. Based on the experiments, it is empirically suggested to adjust the tradeoff parameter from $5\times 10^{-5}$ to $5\times 10^{-4}$.

\begin{figure*}[t]\centering
  \includegraphics[width=0.98\textwidth]{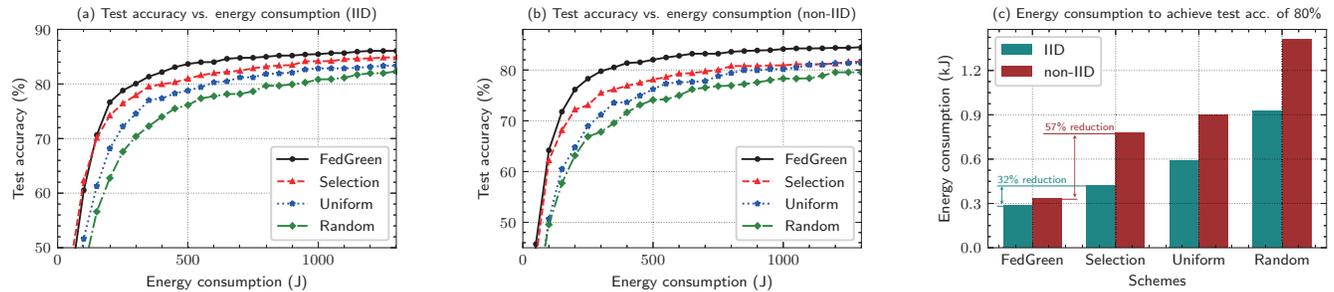}
  \caption{Performance comparison with various baseline schemes on CIFAR-10 dataset.}\label{perform}
  \vspace{-0.4cm}
\end{figure*}

{\bf Comparison with baseline}. Next, we compare our scheme FedGreen with the following baseline schemes.
\begin{itemize}
	\item \textbf{Random}. Each device utilizes a random strategy of gradient compression and $\alpha _i$ is randomly selected from $\left[50, 300\right]$. Computing frequency $f_i$ is calculated according to Eqn.~(\ref{optf}).
	
	\item \textbf{Uniform}. All the devices utilize  an identical  ratio for gradient compression \cite{8889996}. We calculate the average compression ratio $\overline{\alpha}$ of FedGreen, and obtain $\{\alpha _i = \overline{\alpha}, \forall i\}$.
	
	\item \textbf{Selection}. Motivated by \cite{nishio2019client}, we exclude the top 25\% of the devices with the largest energy consumption in the uniform policy.
\end{itemize}
The convergence curves of these schemes over the consumed energy consumption under the IID and non-IID setting are shown as Figs.~\ref{perform}(a) and \ref{perform}(b), respectively. With the same energy consumption requirement, our scheme outperforms the existing schemes to improve the global model accuracy. 
Meanwhile, as the record of our experiments, FedGreen achieves the best final test accuracy in both IID and non-IID settings.
In addition, we provide the experiment results of required energy consumption for achieving 80\% test accuracy in Fig.~\ref{perform}(c). We realize that FedGreen is indeed superior to the above baseline schemes, which consumes the least energy for the convergence performance. Particularly, to achieve the same test accuracy of 80\%, compared with the Selection scheme,  FedGreen reduces 32\% and 57\% of the energy consumption under the IID and non-IID setting, respectively.

\section{Conclusion}
We introduce fine-grained gradient compression for FL in MEC, and proposes FedGreen to dynamically adjust the compression ratios of the devices in an energy-efficient way. We present the basic operations to enable different devices to adopt different compression ratios on demand. Furthermore, we pay attention to the overall performance of FL with gradient compression and study a learning accuracy-energy efficiency tradeoff problem. Based on the applicable methods, we find the approximately optimal compression ratio and computing frequency for each device. Numerical experiments demonstrate that our scheme outperforms the baseline schemes in saving energy on the device side while guaranteeing the accuracy performance of FL in MEC.

\section*{Acknowledgment}
Rong Yu is the corresponding author of this paper. The work is supported in part by National Natural Science Foundation of China (No. 61971148, No. 62001125), Guangxi Natural Science Foundation, China (No. 2018GXNSFDA281013), and Foundation for Science and Technology Project of Guilin City (No. 20190214-3). The work of M. Pan was supported in part by the U.S. National Science Foundation under grants CNS-1801925, CNS-2029569, and CNS 2107057.

\bibliographystyle{IEEEtran}
\bibliography{reference.bib}

\end{document}